\documentclass[letterpaper, 10 pt, conference]{ieeeconf}  
\IEEEoverridecommandlockouts                              
\usepackage{amsmath,amsfonts,amssymb}
\usepackage{prettyref}
\usepackage{mathrsfs}
\usepackage{graphicx}
\usepackage{wrapfig}
\usepackage{xcolor}
\usepackage{subfig}
\usepackage{MnSymbol}
\usepackage{multirow}
\usepackage{booktabs}
\usepackage{algorithm}
\usepackage{sidecap}
\sidecaptionvpos{figure}{c}
\usepackage[noend]{algpseudocode}
\usepackage
[backend=bibtex,
bibstyle=ieee,
citestyle=numeric,
mincitenames=1,
maxcitenames=2,
natbib=true,
doi=false,
isbn=false,
url=false,
eprint=false]{biblatex}
\usepackage[colorlinks,allcolors=gray,hypertexnames=true]{hyperref} 
\newcommand{\citewithauthor}[1]{\citeauthor{#1} \cite{#1}}
\usepackage{soul}
\usepackage{diagbox}
\usepackage{slashbox}

\newtheorem{theorem}{Theorem}[section]
\newtheorem{lemma}[theorem]{\TE{Lemma}}

\algnewcommand{\LineComment}[1]{\State \(\triangleright\) #1}
\algdef{SE}[DOWHILE]{Do}{doWhile}{\algorithmicdo}[1]{\algorithmicwhile\ #1}
\newcommand*{\colorboxed}{}
\def\colorboxed#1#{%
  \colorboxedAux{#1}%
}
\newcommand*{\colorboxedAux}[3]{%
  \begingroup
    \colorlet{cb@saved}{.}%
    \color#1{#2}%
    \boxed{%
      \color{cb@saved}%
      #3%
    }%
  \endgroup
}

\newrefformat{Fig}{Figure~\ref{#1}}
\newrefformat{fig}{Figure~\ref{#1}}
\newrefformat{par}{Section~\ref{#1}}
\newrefformat{appen}{Appendix~\ref{#1}}
\newrefformat{sec}{Section~\ref{#1}}
\newrefformat{sub}{Section~\ref{#1}}
\newrefformat{table}{Table~\ref{#1}}
\newrefformat{alg}{Algorithm~\ref{#1}}
\newrefformat{Alg}{Algorithm~\ref{#1}}
\newrefformat{Def}{Definition~\ref{#1}}
\newrefformat{Thm}{Theorem~\ref{#1}}
\newrefformat{Lem}{Lemma~\ref{#1}}
\newrefformat{step}{Step~\ref{#1}}
\newrefformat{ln}{Line~\ref{#1}}
\newrefformat{eq}{Equation~\ref{#1}}
\newrefformat{pb}{Problem~\ref{#1}}
\newrefformat{it}{Item~\ref{#1}}
\newrefformat{te}{Term~\ref{#1}}
\def\Eqref Eq:#1:{\eqref{eq:#1}}
\newrefformat{Eq}{Equation~\Eqref#1:}

\newcommand{\E}[1]{\mathbf{#1}}
\newcommand{\TE}[1]{\textbf{#1}}

\newcommand{\TWO}[2]{\left(\setlength{\arraycolsep}{1pt}\begin{array}{cc}{#1} & {#2}\end{array}\right)}

\newcommand{\MTT}[4]{\left(\setlength{\arraycolsep}{1pt}\begin{array}{cc}#1 & #2 \\ #3 & #4\end{array}\right)}

\newcommand{\fmin}[1]{\underset{#1}{\min}}
\newcommand{\fmax}[1]{\underset{#1}{\max}}
\newcommand{\argmin}[1]{\underset{#1}{\text{argmin}}}
\newcommand{\argminP}[1]{\text{argmin}}

\definecolor{XXXRED}{RGB}{219,52,57}

\newcommand{\WORK}{\mathcal{W}}
\newcommand{\RR}{\mathbb{R}}
\newcommand{\POSS}{\mathcal{X}}
\newcommand{\DPOSS}{\bar{\POSS}}
\newcommand{\POS}{\E{x}}
\newcommand{\DPOS}{\bar{\POS}}
\newcommand{\GRAPH}{\mathcal{G}}
\newcommand{\DGRAPH}{\bar{\GRAPH}}
\newcommand{\EE}{\mathcal{E}}
\newcommand{\DEE}{\bar{\EE}}
\newcommand{\ee}{\E{e}}
\newcommand{\dee}{\bar{\ee}}
\newcommand{\METRIC}{\mathcal{M}}
\newcommand{\MAPPING}{\Phi}

\newcommand{\proofread}[1]{}
\newif\ifArxiv

\AtBeginBibliography{\footnotesize}
\makeatletter
\IEEEtriggercmd{\reset@font\normalfont\fontsize{7.0pt}{7.2pt}\selectfont}
\makeatother
\IEEEtriggeratref{1}

\makeatletter
\newcommand\fs@ruled@notop{\def\@fs@cfont{\bfseries}\let\@fs@capt\floatc@ruled
  \def\@fs@pre{}%
  \def\@fs@post{\kern2pt\hrule\relax}%
  \def\@fs@mid{\kern2pt\hrule\kern2pt}%
  \let\@fs@iftopcapt\iftrue}
\renewcommand\fst@algorithm{\fs@ruled@notop}
\makeatother

\title{\large\bf Multi-Robot Path Planning in Complex Environments via Graph Embedding \vspace{-10px}}
\author{Xifeng Gao$^{1}$, Zherong Pan$^{2}$, and Ruiqi Ni$^{1}$ \vspace{-20px}  \\
\thanks{ $^{1}$Xifeng Gao and Ruiqi Ni are with Department of Computer Science, Florida State University (gao@cs.fsu.edu and rn19g@my.fsu.edu). $^2$Zherong Pan is with Department of Computer Science, University of Illinois at Urbana-Champaign (zherong@illinois.edu).}}
        
\IEEEaftertitletext{\vspace{-0.5\baselineskip}}
\begin{document}
\maketitle
\thispagestyle{empty}
\pagestyle{empty}

\begin{abstract}
We propose an approach to solve multi-agent path planning (MPP) problems for complex environments. Our method first designs a special pebble graph with a set of feasibility constraints, under which MPP problems have feasibility guarantee. We further propose an algorithm to greedily improve the optimality of planned MPP solutions via parallel pebble motions. As a second step, we develop a mesh optimization algorithm to embed our pebble graph into arbitrarily complex environments. We show that the feasibility constraints of a pebble graph can be converted into differentiable geometric constraints, such that our mesh optimizer can satisfy these constraints via constrained numerical optimization. We have evaluated the effectiveness and efficiency of our method using a set of environments with complex geometries, on which our method achieves an average of $99.0\%$ free-space coverage and $30.3\%$ robot density within hours of computation on a desktop machine.
\end{abstract}
\section{\label{intro}Introduction}
MPP is a unified theoretical model of various applications, including warehouse arrangement \cite{ChiHanYu2018WAFR}, coverage planning for flooring vacuuming \cite{7989156,6850799}, and inspection planning for search and rescue \cite{Baxter2007}. Some applications, such as flooring vacuuming, search and rescue, require MPP planners to tackle complex environments. Intuitively, a task can be accomplished more efficiently when more robots are available. However, coordinating a large number of robots for complex environments can incur a high computational cost. Indeed, determining the feasibility of simplified MPP instances have been shown to be PSPACE-hard \cite{hopcroft1984complexity} (for rectangular objects) and strongly NP-hard \cite{spirakis1984strong} (for polygonal environments). It is thus unsurprising that brute-force search algorithms \cite{wagner2012probabilistic,doi:10.1177/027836402320556458,Le2017CooperativeMS,sharon2015conflict} are only practical for tens of robots.

\begin{figure}[ht]
\centering
\includegraphics[width=1\linewidth]{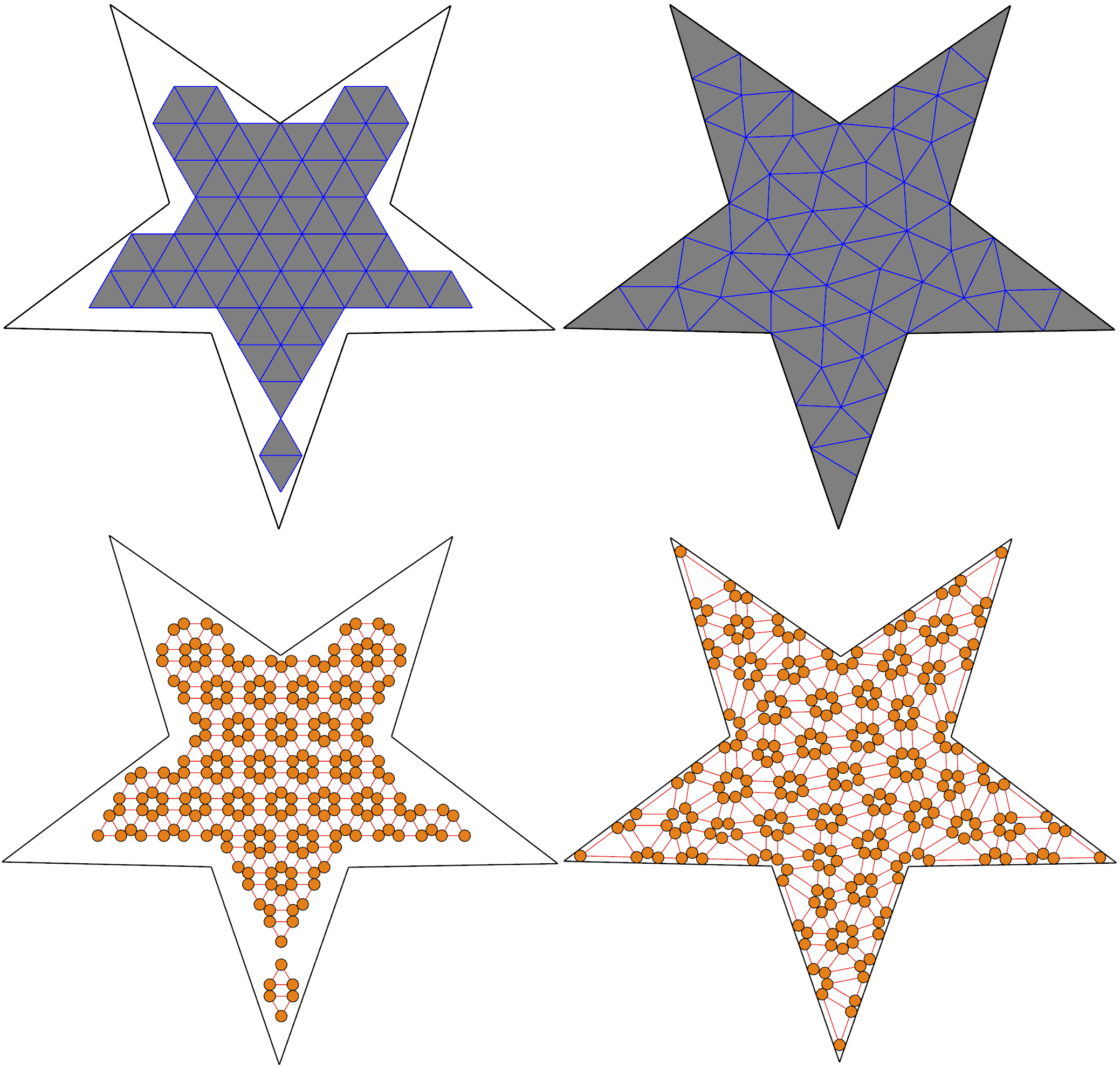}
\caption{\small{\label{fig:irregular} For an environment with non-axis aligned boundaries, compared to a triangulation in the lattice style (top left) and its resulting graph (bottom left), our method generates a boundary aligned triangulation (top right) that can produce a graph with full coverage of the entire environment (bottom right). MPP problems on the graph are always feasible when robots (orange) are arranged on the graph vertices.}}
\vspace{-10px}
\end{figure}

Instead of aiming at generally complex environments, a row of prior works \cite{auletta1999linear,110,yu2015pebble} assume that robots can only move on a discrete graph, i.e. robots reside in a discrete set of vertices and move along a discrete set of paths connecting the vertices. For special graphs, such as trees \cite{auletta1999linear,yu2015pebble} and regular grids \cite{conf/iros/HanRY18,Yu19AR}, determining feasibility and even finding near optimal solutions are polynomial-time solvable. In his recent work, \citewithauthor{Yu19AR} proposed to use curved grids to tile complex environments, but no practical algorithm has been proposed to construct such grids for general geometries. When the shape of an environment has inherent irregularity as illustrated in \prettyref{fig:irregular}, regular geometric patterns can generate a low-level of coverage or miss curved-boundary regions.

\TE{Main Results:} We propose a MPP planner that generalizes the idea of graph-based algorithms \cite{ChiHanYu2018WAFR,YuLav16TOR,Yu19AR,110} to complex environments with curved boundaries. Our method uses a special graph topology, on which robots can both move along edges and spin in loops. When the graph satisfy certain geometric properties, it has been shown in \cite{2002.11892,yu2015pebble} that MPP instances are always feasible under certain geometric conditions. We further propose a space-time search algorithm to find parallel motions and greedily improve the makespan. As our key contribution, we show that our graph can be topologically identified with a planar triangle mesh, where the geometric properties of the graph are mapped to differentiable geometric constraints of the triangles. As a result, we can adapt the greedy triangle mesh optimization framework proposed in \cite{hoppe1999new,hoppe1993mesh} to optimize a graph that can be embedded into arbitrarily complex environments. During the mesh optimization, a constrained numerical optimization is used to ensure all the geometric constraints are satisfied, i.e., MPP instances are feasible on the graph.

We have evaluated our method on a set of $15$ complex environments including 2D floorplans of houses, malls, and maps. Our algorithm is efficient, allowing a graph containing thousands of robots to be generated within an hour on average using a desktop machine. Our generated graph embedding not only aligns well with the curved input boundary but also has high free-space coverage ratio of $99.0\%$ and robot density ratio of $30.3\%$ on average. In addition, compared to sequential scheduling, our space-time search algorithm empirically reduces the makespan by a factor of $15\times$.
\section{\label{sec:related}Related Work}
In this section, we review related works on MPP, mesh optimization, and graph embedding.

\TE{MPP} problems in a general setting is intractable \cite{hopcroft1984complexity,spirakis1984strong} and practical methods target at solving restricted problem subclass. If the number of robots is small, then brute-force search is tractable. Early works \cite{doi:10.1177/027836402320556458,inproceedingsrss05} use RRT or PRM as sub-problem solvers and are thus probabilistically complete. When robots are restricted to move on a graph, their motions can be enumerated using conflict-based search \cite{sharon2015conflict} or integer programming \cite{YuLav16TOR}. These algorithms are deterministically complete and optimal. For larger robot swarms, state-of-the-art results restrict robots to a graph of special topology such as trees \cite{auletta1999linear} and loops \cite{yu2015pebble}, where feasible solutions can be found in polynomial time, and regular grids \cite{Yu19AR}, where near optimal solutions can be found up to a constant factor. Our method resembles \cite{yu2015pebble} in that we allow robots to move and spin. To adapt our method to arbitrarily complex geometries, we cannot guarantee optimality, but we greedily improve optimality via space-time search. 

\TE{Mesh Optimization} algorithm was originally proposed in \cite{hoppe1993mesh}. The algorithm defines a set of possible remeshing operators that change both the geometry and topology of a mesh and exhaustively applies these operators to monotonically improve a given quality metric. Mesh optimization finds applications in geometric processing \cite{hoppe1999new,alliez2008recent}, and physics-based modeling \cite{Feldman:2005:AGH,villard2005adaptive}. These applications use similar algorithm frameworks but differ in the set of operators and the quality metric. Although the framework proposed in \cite{hoppe1993mesh} is greedy and only achieves local optimality, it is capable of exploring a large portion of solution space. In robotic research, there are similar ideas of cell decomposition \cite{lingelbach2005path,latombe1991exact} and resolution-complete motion planning. Meshes are also used in distributed multi-robot coordination \cite{8594302,lee2009adaptive} that considers moving a group of robots while maintaining given topological patterns.

\TE{Graph Embedding} applies the discrete graph theory of MPP to continuous environments. It has been used in \cite{conf/iros/HanRY18,ChiHanYu2018WAFR,Yu19AR} that consider rectangular, triangular, and hexagonal grids. These works do not allow the graph topology or the geometry to be modified. Optimal graph embedding problems have also been applied to network optimization and embedding \cite{mutzel2002bend,fischer2013virtual}, where the topology of the graph can be modified but the geometries (vertex positions of the graph) are fixed. Recent work on layout optimization \cite{peng2014computing,10.1145/3197517.3201306} allows simultaneous modification of geometry and topology for urban city designing. Graph embedding is considered more challenging than mesh optimization because it has to preserve a set of topological constraints. We show that the special topological constraints in MPP problems can be preserved during mesh optimization by limiting the set of legal remeshing operators in \cite{hoppe1993mesh}.
\section{\label{sec:problem}Problem Statement}
We consider planar, labeled, disk-shaped MPP in a complex workspace $\WORK\subset\RR^2$ with piecewise linear boundaries. We assume there are $N$ robots centered at $\POS_{1,\cdots,N}\in\WORK$. All the robots have an identical radius $r$, so the $i$th robot takes up the circular region: $C(\POS_i)\triangleq\{\POS|\|\POS-\POS_i\|\leq r\}\subset\WORK$. Our MPP problem is defined as a permutation $\sigma$ of robot locations, such that the start position of the $i$th robot is $\POS_i$ and its goal position is $\POS_{\sigma(i)}$. As a result, our robots can only move to finitely many positions, we further restrict the robots' moving paths to a finite set of edges. In other words, robots are restricted to a graph $\GRAPH=<\POSS,\EE>$ that can be embedded into $\WORK$. Here $\POSS$ is the set of vertices mapped to robots' positions $\POSS\triangleq\{\POS_1,\cdots,\POS_N\}$ and we refer to a graph vertex and a robot position interchangeably. $\EE$ is the set of edges connecting vertices in $\POSS$ and we refer to a path connecting two positions in $\WORK$ and a graph edge interchangeably.

Our method consists of two steps. In our first step (\prettyref{sec:mpp}), we consider a subclass of graph topology that is a special case of \cite{2002.11892}. Our graph consists of simple loops with 3 vertices. These 3-loops are connected by extra edges into a single, connected components. Two kinds of moves are available for the robot: A \TE{cyclic move} allows a group of 3 robots to cyclically permute locations along a 3-loop. A \TE{vacant move} allows a robot to move along any edge $\ee\in\EE$ connecting $\POS_i$ and $\POS_j$, as long as $\POS_j$ is not occupied by any other robot (a vacant vertex). It has been shown in \cite{2002.11892} that MPP problems are feasible on such graphs, as long as the number of robots is less than $|\POSS|$. However, if the number of robots is strictly $|\POSS|-1$, then the worst case makespan can be $\mathcal{O}(|\POSS|^2)$. As the number of robots decreases, more vacant vertices are available that allows multiple robots to move simultaneous. To exploit extra vacant vertices, we present a space-time search algorithm that schedule simultaneous robot motions to improve the makespan.

While our first step only considers the graph topology, our second step (\prettyref{sec:graph2mesh} and \prettyref{sec:meshopt}) focuses on the geometry of the graph. Specifically, we compute a planar embedded of $\GRAPH$ into $\WORK$ such that, for any $\POS_i\in\POSS$, $C(\POS_i)\subset\WORK$ and $C(\POS_i)\cap C(\POS_j)=\emptyset$ for any $i\neq j$. We further ensure that a cyclic move on any 3-loop and a vacant move along any $\ee\in\EE$ can be performed in a collision-free manner. A planar embedding satisfying the above conditions are considered feasible. In addition to ensuring feasibility, our algorithm maximizes a user-defined metric function $\METRIC(\GRAPH)$ that encodes different requirements for a ``good'' graph. 
\section{\label{sec:mpp} MPP on Loop Graphs}
In this section, we assume that robots can move on $\GRAPH$ in a collision-free manner and present the graph's topological properties, under which any MPP problems can be solved. 

\begin{figure}[ht]
\centering
\scalebox{0.8}{
\includegraphics[width=1\linewidth]{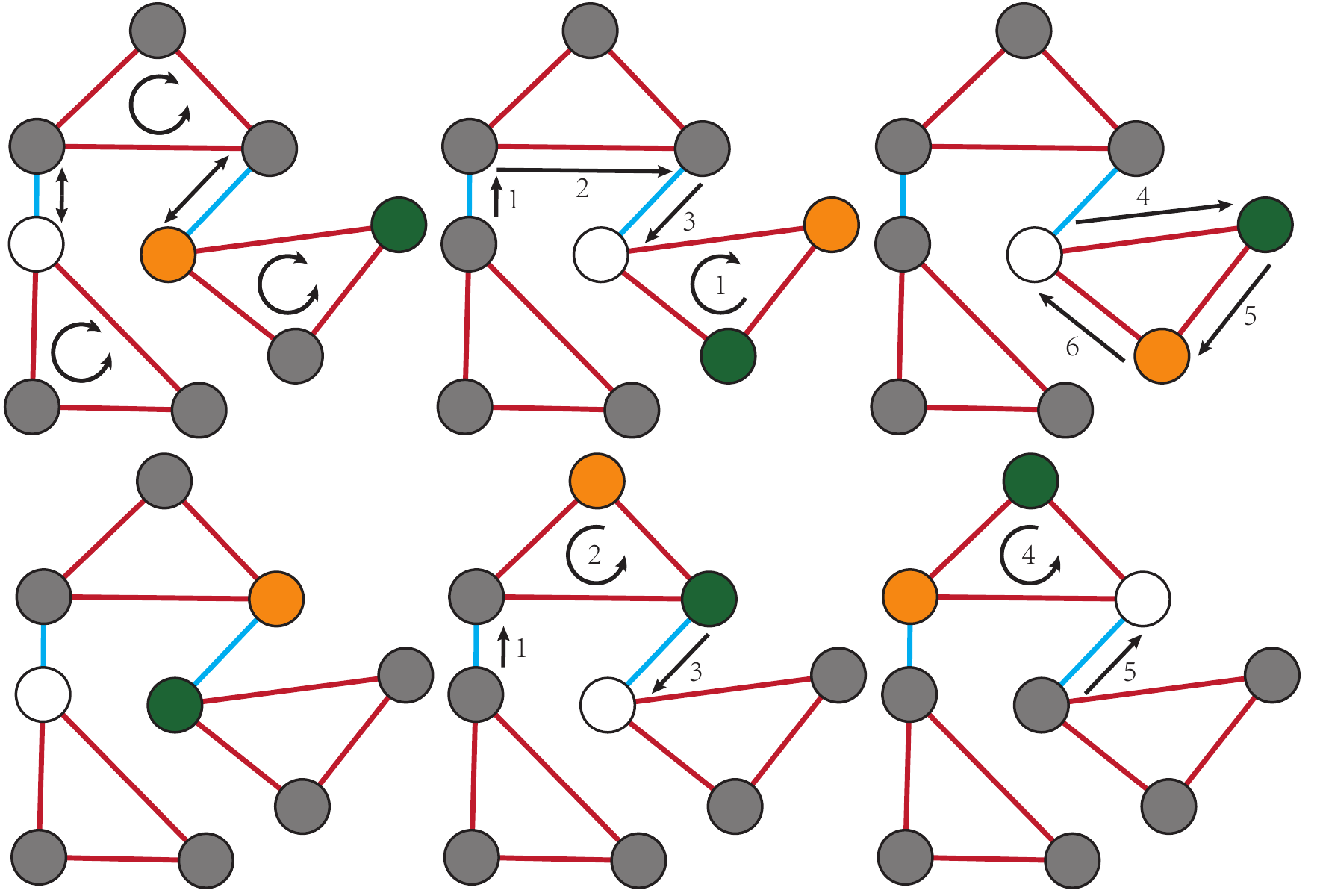}
\put(-200,155){\TE{Case I}}
\put(-200,70){\TE{Case II}}}
\caption{\small{\label{fig:topology} Our graph (a) consists of 3-loops (red edges), along which cyclic moves can be performed along (black, arced) arrows, connected by (blue) inter-loop edges, along which vacant moves can be performed along (black) arrows. An MPP problem can be solved using a series of sub-swaps between two connected vertices. \TE{Case I:} If the two vertices, blue and yellow, belong to the same loop, then we can move the vacant vertex to the loop (1,2) and swap the two vertices via 3 cyclic moves (4,5,6). \TE{Case II:} If two vertices belong to two connected loops, then we can use (1,2,3,4) to reduce to \TE{Case I}. After the swaps, all other robots' position are unaltered using a reversed sequence of moves.}}
\vspace{-10px}
\end{figure}
\subsection{MPP Feasibility}
As illustrated in \prettyref{fig:topology} (a), we consider connected graphs consisting of simple loops with 3 vertices that are connected by additional inter-cell edges. We can show that MPP instances restricted to such graphs are always feasible:
\begin{lemma}
\label{lem:MPPFeasibility}
If $\GRAPH$ has more than 1 loop, then any MPP problem with $N<|\POSS|$ is feasible.
\end{lemma}
We only give a sketch of proof and refer readers to \cite{2002.11892} for a complete proof. Any permutation $\sigma$ of robot positions can be decomposed into a set of pairwise swaps. To swap the location of two robots, $\POS_i$ and $\POS_j$, we first find a path in $\GRAPH$ connecting $\POS_i$ and $\POS_j$, which is always possible as $\GRAPH$ is connected, denoted as $\POS_i,\POS_1,\POS_2,\cdots,\POS_K,\POS_j$.
We can further decompose the swap into a series of sub-swaps:
\begin{align*}
\POS_i\leftrightarrow\POS_1, \cdots, \POS_i\leftrightarrow\POS_K, \POS_i\leftrightarrow\POS_j,
\POS_K\leftrightarrow\POS_j, \cdots, \POS_1\leftrightarrow\POS_j,
\end{align*}
where each pair of positions in a sub-swap are connected directly by some $\ee\in\EE$, where $\ee$ is either a loop edge or an inter-cell edge. In either case, the sub-swap can be performed with the help of a nearby vacant vertex as illustrated in \prettyref{fig:topology}. Finally, such vacant vertex must exist because our number of robots is strictly less than the number of vertices and the vacant position can be moved anywhere via vacant moves. 

\subsection{Parallel Robot Moves}
\begin{wrapfigure}{r}{0.2\textwidth}
\vspace{-25px}
\includegraphics[width=1\linewidth]{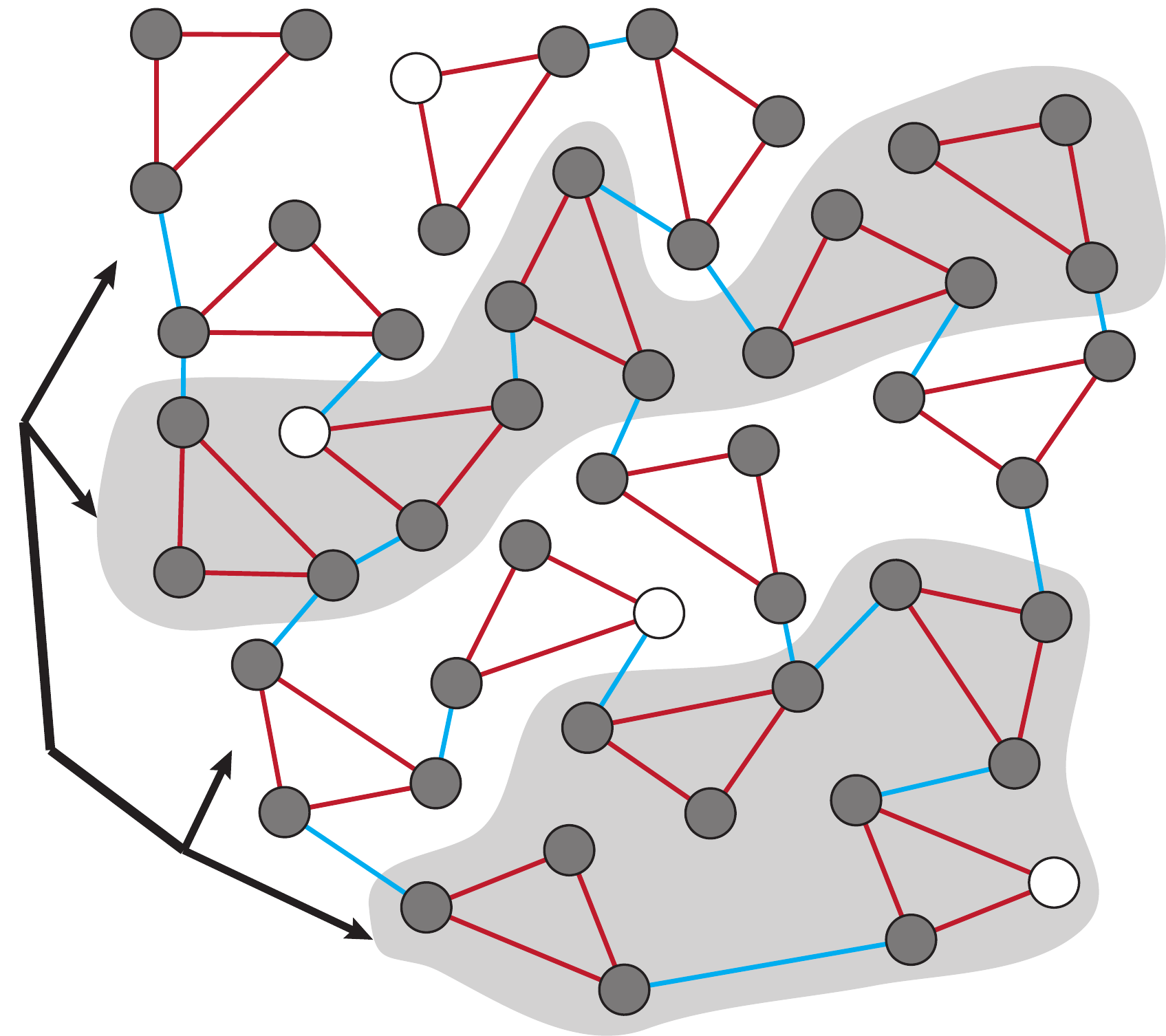}
\put(-100,70){\tiny$\GRAPH_1$}
\put(-96,37){\tiny$\GRAPH_2$}
\put(-92,23){\tiny$\GRAPH_3$}
\put(-70,0){\tiny$\GRAPH_4$}
\vspace{-15px}
\end{wrapfigure}
In the worst case, the makespan of above mentioned MPP solutions is $\mathcal{O}(|\POSS|^2)$. This is because we have only one vacant vertex, which is needed to perform every sub-swap. When more vacant vertices are available, we can parallelize the sub-swaps and improve the makespan. We present a space-time search algorithm to find parallel sub-swaps, which is similar to \cite{Yu19AR} in principle but adapted to graphs of more general topology. As illustrated in the inset, we first use the CLINK algorithm \cite{defays1977efficient} to cluster the loops into a binary tree. This algorithm starts by treating each loop as a separate cluster, and then iteratively merge two smallest clusters (with the smallest number of vertices) until only one cluster is left. We then iteratively merge small leaf nodes until all leaf nodes have more than $K>1$ loops ($K=4$ in the inset), where $K$ is a user specified parameter determining the congestion-level. In the worst case, this method requires $\lceil|\POSS|/(3K)\rceil$ vertices to be left vacant. Next, we introduce a vacant vertex for each leaf node. As a result, cyclic moves and vacant moves can be performed parallelly within a leaf node and two pairs of connected leaf nodes can perform two robot position swaps in parallel. Finally, we adopt the divide-and-conquer algorithm proposed in \cite{Yu19AR}. For every internal binary tree node, we consider their two children, which are two connected sub-graph. We swap robots between the two sub-graphs until $\POS_i$ and $\POS_{\sigma(i)}$ belongs to the same sub-graph for all $i=1,\cdots,N$. This procedure is performed recursively from the root node to all the leaves, where the swaps within different sub-trees are performed in parallel.

\subsection{Scheduling Robot Swaps between Sub-Graphs}
The major challenging in the above algorithm lies in the scheduling of parallel robot position swaps between sub-graphs, for which we use a space-time data structure. We denote $\GRAPH_i$ as the $i$th leaf node and we assume there are $N_1$ leaf nodes in one of the sub-graph and $N_2$ in the other ($N_1+N_2\leq\lceil|\POSS|/(3K)\rceil$). The inter-sub-graph position swaps are decomposed into several rounds, where each round consists of several parallel position swaps between two connected leaf nodes. We denote as $\GRAPH_{i,t}$ as the state of $\GRAPH_i$ at the beginning of $t$th round. We further connect $\GRAPH_{i,t}$ and $\GRAPH_{j,t+1}$ using a space-time edge $\ee_{ij,t}$ if two conditions hold: 1) $\GRAPH_i$ and $\GRAPH_j$ are connected by some inter-loop edge $\ee\in\EE$; 2) $\GRAPH_i$ and $\GRAPH_j$ belongs to the same sub-graph, i.e., $i,j\leq N_1$ or $N_1<i,j\leq N_1+N_2$. Our space-time data structure is defined as the space-time graph $<\{\GRAPH_{i,t}|t\leq t_0+1\},\{\ee_{ij,t}|t\leq t_0\}>$, where $t_0$ is the maximal allowed number of rounds. Note that the space-time graph is directed with each $\ee_{ij,t}$ directing from $\ee_{j,t+1}$ to $\ee_{i,t}$. The inter-sub-graph position swaps can be accomplished by multi-rounds of swaps between leaf nodes, which in turn corresponds to selecting a set of space-time edges $\ee_{ij,t}$. Our algorithm greedily select a set of space-time edges leading to an inter-sub-graph position swap in the earliest round. After each selection, we remove the conflicting space-time edges before selecting the next set of edges, until all the robots belong to the correct sub-graph. When no position swaps can be found for the given $t_0$, we introduce a new round and increase $t_0$ by one. This procedure is guaranteed to terminate.

\begin{algorithm}[ht]
\begin{small}
\caption{\label{alg:spacetime} Space-Time Scheduling of Robot Swaps}
\begin{algorithmic}[1]
\Require{Sub-graphs A: leaves $G_{1,\cdots,N_1}$ and $G_{N_1+1,\cdots,N_1+N_2}$}
\State $t_0\gets2$
\State Build space-time graph $<\{\GRAPH_{i,t}|t\leq t_0+1\},\{\ee_{ij,t}|t\leq t_0\}>$
\While{$\sum_{i=1,\cdots,N_1+N_2}\#\GRAPH_{i,t_0}>0$}
\State Found$\gets$False
\For{$t=1,\cdots,t_0$}
\For{Connected leaf nodes $\GRAPH_a,\GRAPH_b$}
\State Run Dijkstra's algorithm from $\GRAPH_{a,t},\GRAPH_{b,t}$
\State until some $\GRAPH_{i,t'}$ satisfying \prettyref{eq:swapout} is found
\If{Found both shortest paths}
\State Select edges on the two shortest paths
\For{Selected edge $\ee_{ij,t}$}
\State Remove $\{\ee_{kj,t},\ee_{ik,t}\}$
\For{$t'=1,\cdots,t_0$}
\State Remove 
\State $\{\ee_{ki,t'},\ee_{kj,t'}|\fmin{t''=t',\cdots,t_0}\#_/\GRAPH_{j,t''}=0\}$
\EndFor
\EndFor
\State Remove $\{\ee_{ak,t},\ee_{bk,t}\}$
\State Found$\gets$True
\EndIf
\EndFor
\EndFor
\If{Found$\gets$False}
\State $t_0\gets t_0+1$
\State Build space-time nodes $\GRAPH_{i,t_0+2}$, edges $\ee_{ij,t_0+1}$
\EndIf
\EndWhile
\end{algorithmic}
\end{small}
\end{algorithm}
Our method is outlined in \prettyref{alg:spacetime}, which is composed of two main steps. First, we consider all pairs of connected leave nodes $<\GRAPH_{a,t},\GRAPH_{b,t}>$ belonging to different sub-graphs. We find the two shortest paths connecting from $\GRAPH_{a,t}$, $\GRAPH_{b,t}$ to some leaf node containing a robot that belongs to a different sub-graph. Specifically, we run Dijkstra's algorithm with all-one edge weights from both $\GRAPH_{a,t}$ and $\GRAPH_{b,t}$. We terminate the Dijkstra's algorithm at any $\GRAPH_{i,t}$ satisfying the following condition:
\begin{equation}
\label{eq:swapout}
\fmin{t'=t,\cdots,t_0}(\#\GRAPH_{i,t'})>0,
\end{equation}
where we define $\#\GRAPH_{i,t}$ as the number of robots within $\GRAPH_{i,t}$ that belongs to a different sub-graph from $\GRAPH_{i,t}$ at the beginning of the $t$th round. Intuitively, \prettyref{eq:swapout} ensures that swapping one more robot away from $\GRAPH_i$ during the $t$th round will not cause conflict with the number of to-be-swapped robots required by other swapping operations in the future rounds. Finally, note that we consider nodes pairs in a time-ascending order. The first pair of leaf nodes, for which such paths can be found, are selected, which corresponds to the position swap in the earliest round. 

Our second step would remove three types of conflicting edges: 1) If an edge $\ee_{ij,t}$ is selected, then all the edges $\{\ee_{kj,t},\ee_{ik,t}|k=1,\cdots,N_1+N_2\}$ are removed because they conflict with the edges along the two paths; 2) all the edges $\{\ee_{ak,t},\ee_{bk,t}|k=1,\cdots,N_1+N_2\}$ are removed because they conflict with the inter-sub-graph swap between $\GRAPH_{a,t}$, $\GRAPH_{b,t}$; 3) Let's define $\#_/\GRAPH_{j,t}$ as the number of robots belonging to the same sub-graph as $\GRAPH_j$ at the beginning of $t$th round ($\#\GRAPH_{j,t}+\#_/\GRAPH_{j,t}+1$ equals to the number of vertices in $\GRAPH_j$), then all the edges: 
\begin{align*}
\{\ee_{kj,t}|\fmin{t'=t,\cdots,t_0}\#_/\GRAPH_{j,t'}=0\land k=1,\cdots,N_1+N_2\},
\end{align*}
should be removed. This is because selecting edge $\ee_{kj,t}$ implies that a robot $\POS_a\in\GRAPH_{k,t}$ must be swapped with another robot $\POS_b\in\GRAPH_{j,t}$ at $t$th round. We must have $\POS_b$ belongs to the same sub-graph as $\GRAPH_j$ at the beginning of $t$th round, because otherwise the shortest path would stop at $\POS_b$ instead of moving on to $\POS_a$. However, $\fmin{t'=t,\cdots,t_0}\#_/\GRAPH_{j,t'}=0$ implies that, if $\POS_b$ was swapped out of $\GRAPH_j$ during $t$th round, there will not be enough to-be-swapped agents (that belongs to the same sub-graph as $\GRAPH_j$) during some future rounds.
\section{\label{sec:graph2mesh}Graph Embedding and Mesh Discretization}
\begin{figure}[ht]
\centering
\includegraphics[width=0.99\linewidth]{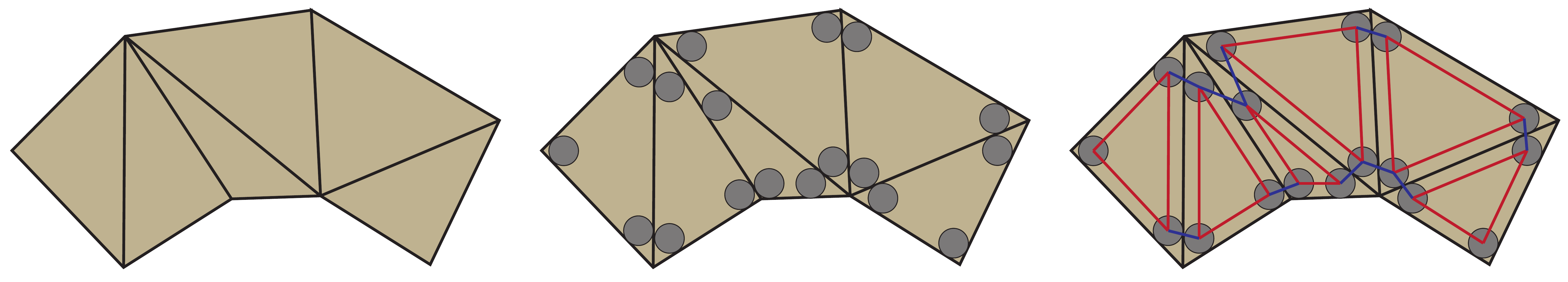}
\put(-205,1){(a)}
\put(-123,1){(b)}
\put(-40,1){(c)}
\caption{\small{\label{fig:triangular_embedding} (a): A mesh discretizing $\WORK$ is another graph $\DGRAPH$ that is also a cell decomposition with straight-line edges and triangular cells. (b): We map $\DGRAPH$ to $\GRAPH$ by first putting robots (gray) on the cell's corner points. (c): We then create loop edges for robots in a single cell (red), and finally create inter-cell edges between robots of neighboring cells (blue).}}
\vspace{-10px}
\end{figure}

In this section, we move on to consider the geometric properties of $\GRAPH$ that allows robots to move in a collision-free manner within a $\WORK$ of arbitraily complex geometry. Our key innovation is a mapping from a mesh discretizing $\WORK$ to a graph embedded in $\WORK$. As illustrated in \prettyref{fig:triangular_embedding} (a), a mesh discretizing $\WORK$, denoted as $\DGRAPH=<\DPOSS,\DEE>$, is another graph that is also a simplicial-complex, cell-decomposition of $\WORK$ and we denote all the variables defined on the mesh using a bar over variables. Given a mesh $\DGRAPH$, we can convert it into a graph using \prettyref{alg:graph2mesh}, which is denoted as a mapping $\MAPPING(\DGRAPH)=\GRAPH$. \prettyref{alg:graph2mesh} consists of 4 general steps. First, we place robots on inner corner points of each cell, i.e. points inside a cell that are distance-$r$ away from two consecutive edges of the cell. Next, we assume that 3 robots inside the same cell form a loop, along which cyclic moves can be performed, and we add 3 loop edges to $\GRAPH$. We then add inter-cell edges to $\GRAPH$ between the two pairs of corner points sharing an edge $\dee$ in $\DGRAPH$, along which vacant move can be performed. Note that some cells may be too small and robots placed on inner corners are not collision-free. Even when corner points are collision-free, robots inside the same cell can still collide when performing cyclic moves. Therefore, we add a fourth and final step to remove from $\GRAPH$ all the corner points (and incident edges) in invalid cells. We define a cell as invalid if the 3 corner points or cyclic moves are not collision-free. In the following, we show that the collision-free conditions of $\GRAPH$ can be converted to three differentiable geometric constraints on $\DGRAPH$.
\begin{algorithm}[ht]
\begin{small}
\caption{\label{alg:graph2mesh} Evaluate $\MAPPING(\DGRAPH)$.}
\begin{algorithmic}[1]
\State $\POSS\gets\emptyset$ and $\EE\gets\emptyset$
\For{Each triangular cell in $\DGRAPH$}
\For{Corner point $\POS_{1,2,3}$ in the cell}
\State $\POSS\gets\POSS\bigcup\{\POS_i\}$
\EndFor
\For{$\ee$ between two points in $\POS_{1,2,3}$}
\State $\EE\gets\EE\bigcup\{\ee\}$
\EndFor
\EndFor
\For{$\ee$ between neighboring corner points sharing edge in $\DGRAPH$}
\State $\EE\gets\EE\bigcup\{\ee\}$
\EndFor
\State $\GRAPH\gets<\POSS,\EE>$
\For{Corner points $\POS$ of invalid cell}
\State Remove $\POS$ from $\GRAPH$
\EndFor
\State Return $\GRAPH$
\end{algorithmic}
\end{small}
\end{algorithm}

\subsection{Condition 1: Planner Embedding}
Since $\DGRAPH$ is a simplicial complex, each cell is a triangle and we denote the 3 vertices of this triangle as: $\DPOS_{1,2,3}$. The distance-$r$ corner points can be computed using the following formula, as illustrated in \prettyref{fig:invalid} (a):
\begin{align}
\label{eq:corner}
\POS_2 =\DPOS_2+r
\frac{(\DPOS_3-\DPOS_2)\|\DPOS_1-\DPOS_2\|+(\DPOS_1-\DPOS_2)\|\DPOS_3-\DPOS_2\|}
{\|(\DPOS_3-\DPOS_2)\times(\DPOS_1-\DPOS_2)\|}.
\end{align}
We only show formula for $\POS_2$ and the formulas for $\POS_{1,3}$ are symmetric. We will mark the cell as invalid and remove the 3 vertices from $\GRAPH$ if there are overlappings between $C(\POS_{1,2,3})$. 

\subsection{Condition 2: Collision-Free Cyclic Moves}
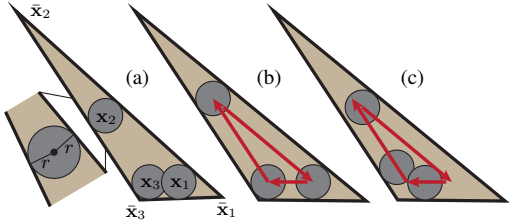
\begin{figure}[ht]
\centering
\vspace{-15px}
\scalebox{0.8}{
\includegraphics[width=0.99\linewidth]{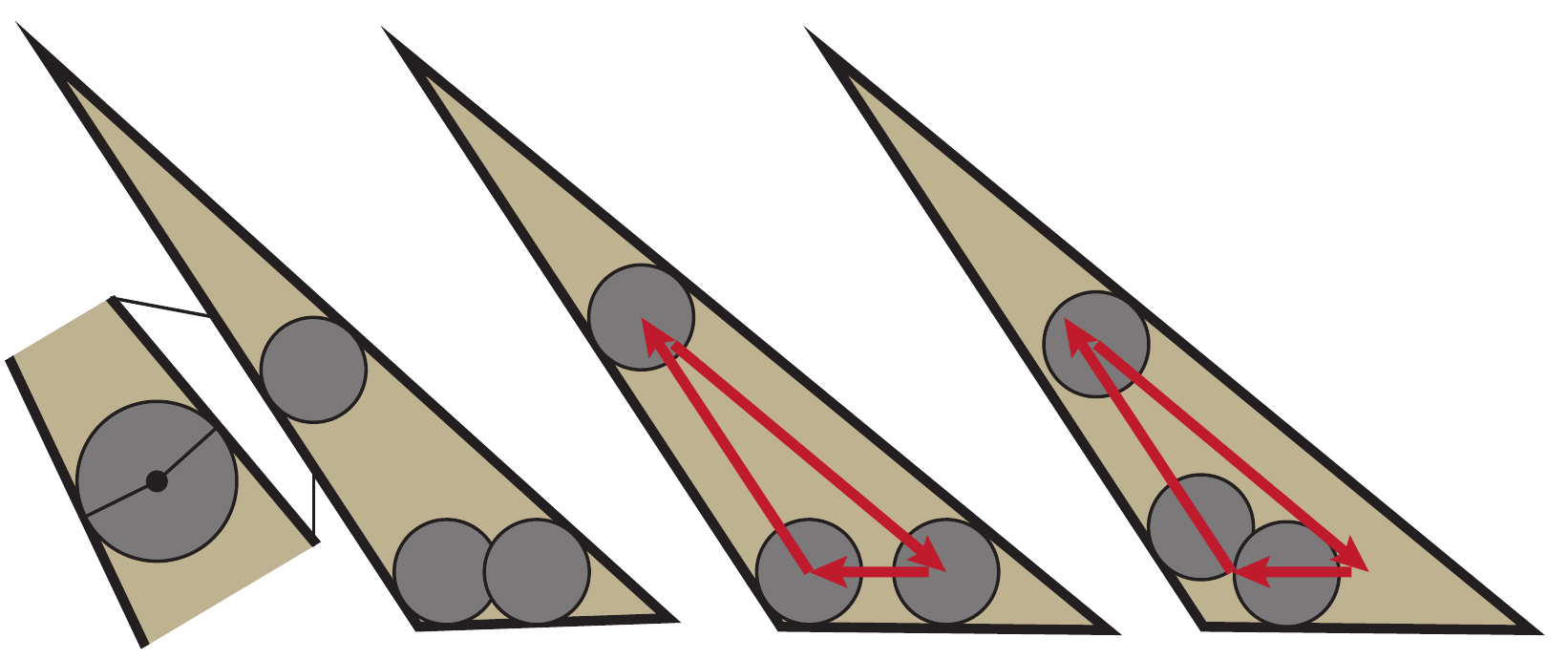}
\put(-225,20){\small{$r$}}
\put(-214,27){\small{$r$}}
\put(-142,-2){\small{$\bar{\POS}_1$}}
\put(-230,92){\small{$\bar{\POS}_2$}}
\put(-185,-3){\small{$\bar{\POS}_3$}}
\put(-164,11){\small{$\POS_1$}}
\put(-199,42){\small{$\POS_2$}}
\put(-178,11){\small{$\POS_3$}}
\put(-185,60){(a)}
\put(-123,60){(b)}
\put(-55 ,60){(c)}}
\caption{\small{\label{fig:invalid} (a): For a thin triangle, corner points will not be collision-free, violating condition 1. (b): Even when condition 1 holds, robots need to follow the red arrows and perform a cyclic move. (c): If collisions happen during a cyclic move, condition 2 is violated.}}
\vspace{-10px}
\end{figure}

Even when the 3 corner points are not overlapping, robots can still collide when they perform cyclic moves along the 3 loop edges with constant speed, as illustrated in \prettyref{fig:invalid} (b). 
To derive a condition for collision-free cyclic moves, we assume that the 3 robots trace out a trajectory $\tau_{1,2,3}(t)$ where $t\in[0,1]$. At time $t$, the 3 robots are at positions:
\begin{align*}
\POS_{1,2,3}(t)&\triangleq\POS_{1,2,3}(1-t)+\POS_{2,3,1}t,
\end{align*}
and we have collision-free cyclic moves if:
\begin{align}
\label{eq:cond2}
\|\POS_{1,2,3}(t)-\POS_{2,3,1}(t)\|\geq 2r\quad \forall t\in[0,1].
\end{align}
Here we use cyclic subscripts to denote 3 symmetric conditions. The left-hand side of \prettyref{eq:cond2} is quadratic when squared and determining the smallest value of a quadratic equation in $[0,1]$ has closed-form solution, which can be used for determining whether a cell is invalid in \prettyref{alg:graph2mesh}. 

In addition to determining the validity of a cell, our mesh optimization algorithm (\prettyref{sec:meshopt}) requires an operator that can modify an invalid cell's geometric shape to achieve validity using numerical optimization. To this end, we derive a condition equivalent to \prettyref{eq:cond2} but does not contain continuous time variable $t$, because the variable $t$ can take infinitely many values from $[0,1]$ leading to a difficult semi-infinite programming problem \cite{sinha2017review}. Taking one of the equations $\|\POS_1(t)-\POS_2(t)\|\geq 2r$ in \prettyref{eq:cond2} for example (the other 2 cases are symmetric), its left-hand-side is a polynomial of a single time variable $t$. To eliminate $t$, the Fekete, Markov-Luka\'cz theorem \cite{laurent2009sums} can be applied to show that, if \prettyref{eq:cond2} holds, then we have:
\small
\begin{align*}
&\|\POS_1(t)-\POS_2(t)\|^2-4r^2=
\alpha_1(2t-1)^2+2\alpha_2\alpha+\alpha_3+\alpha_4(4t-4t^2) \\
&\land\MTT{\alpha_1}{\alpha_2}{\alpha_2}{\alpha_3}\succeq0\land\alpha_4\geq0,
\end{align*}
\normalsize
where $\alpha_{1,2,3,4}$ are four unknown variables to be fitted. The $2\times2$ PSD-cone constraint is equivalent to two quadratic constraints: $\alpha_1\alpha_3\geq0$ and $\alpha_1\alpha_3\geq\alpha_2^2$. By equating coefficients in the quadratic constraints, we can express $\alpha_{1,2,3,4}$ in terms of $\POS_{1,2,3}$ to get the following equivalent form:
\small
\begin{equation}
\begin{aligned}
\label{eq:equi_cond2}
&(\frac{1}{4}\|2\POS_2-\POS_1-\POS_3\|^2+\alpha_4)(\frac{1}{4}\|\POS_1-\POS_3\|^2-4r^2-\alpha_4)\geq    \\
&\left[\frac{1}{4}(\POS_1-\POS_3)(2\POS_2-\POS_1-\POS_3)\right]^2\land\alpha_4\geq0,
\end{aligned}
\end{equation}
\normalsize
where $\alpha_4$ is an additional decision variable that cannot be eliminated. But unlike $t$, we only need to satisfy \prettyref{eq:equi_cond2} for a single $\alpha_4$. We summarize this result below:
\begin{lemma}
\label{lem:condition2}
A cell in $\DGRAPH$ is valid if its 3 corner points, computed according to \prettyref{eq:corner}, satisfy \prettyref{eq:cond2} for all $t\in[0,1]$ or if the 3 corner points satisfy \prettyref{eq:equi_cond2} for any positive $\alpha_4$.
\end{lemma}

\subsection{Condition 3: Collision-Free Vacant Moves}
We show that, as long as condition 2 is satisfied, condition 3 must also be satisfied, so $\MAPPING(\DGRAPH)$ always satisfy condition 3. Condition 3 requires that a robot can move to a vacant position along any $\ee\in\EE$. We analyze two cases: $\ee$ being a loop edge or $\ee$ being an inter-cell edge.

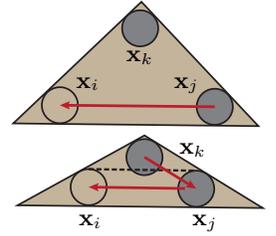
\begin{wrapfigure}{r}{0.2\textwidth}
\vspace{-10px}
\includegraphics[width=0.2\textwidth]{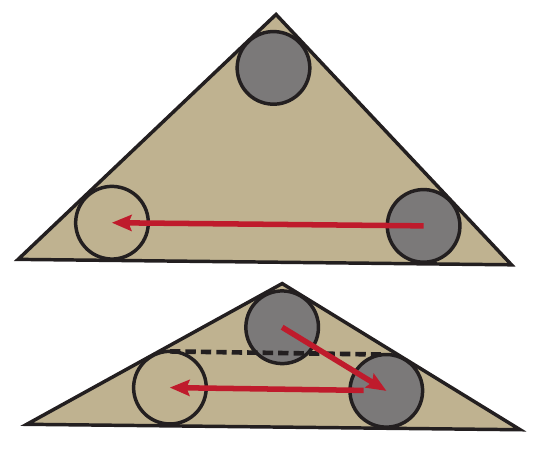}
\put(-74,50){\small{$\POS_i$}}
\put(-37,50){\small{$\POS_j$}}
\put(-55,60){\small{$\POS_k$}}
\put(-73,-2){\small{$\POS_i$}}
\put(-30,-2){\small{$\POS_j$}}
\put(-35,25){\small{$\POS_k$}}
\caption{\small{\label{fig:vacant_loop} Top: the path between $\POS_i$ and $\POS_j$ will not be blocked by $\POS_k$. Bottom: If $\POS_k$ is below the dashed line and blocking the path, then the cyclic move from $\POS_k$ to $\POS_j$ and $\POS_j$ to $\POS_i$ along the red arrows will not be collision-free, contradicting condition 2.}}
\vspace{-5px}
\end{wrapfigure}
\TE{Loop edge:} If one of the 3 corner points $\POS_i$ in a cell is vacant and another robot $\POS_j$ is moving along a loop edge to $\POS_i$, we show that this move is always collision-free. We prove by contradiction as illustrated in \prettyref{fig:vacant_loop}. If this move is not collision-free, then the third corner point $\POS_k$ must be in the way between $\POS_i$ and $\POS_j$. However, in this case, cyclic moves in the cell is not collision-free, which contradicts condition 2.

\begin{figure}[ht]
\centering
\vspace{-5px}
\includegraphics[width=0.99\linewidth]{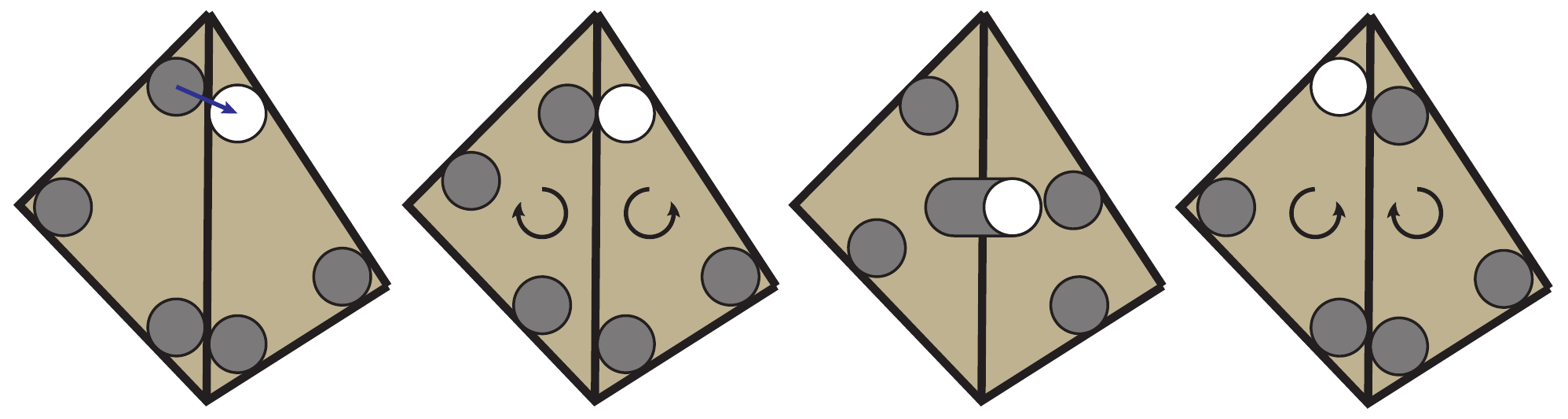}
\put(-224,42){\small{$\POS_i$}}
\put(-207,37){\small{$\POS_j$}}
\put(-219,27){\small{$\dee$}}
\put(-200,50){(a)}
\put(-140,50){(b)}
\put(-80 ,50){(c)}
\put(-20 ,50){(d)}
\caption{\small{\label{fig:vacant_inter_cell} If $\POS_i$ is to be moved to a vacant position $\POS_j$ along a blue inter-cell edge (a), then we can rotate $\POS_i$ clockwise and $\POS_j$ counterclockwise to align $\POS_i$ and $\POS_j$ (b), so that the convex hull of $C(\POS_i)$ and $C(\POS_j)$ does not contain other agents (c). After the vacant move inside the convex hull, we rotate $\POS_{i,j}$ reversely to undo extra changes (d).}}
\vspace{-10px}
\end{figure}
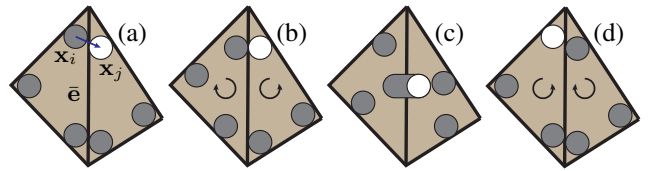

\TE{Inter-cell edge:} If two neighboring triangles satisfy condition 2 and share an edge $\dee\in\DEE$, then cyclic moves inside each triangle are collision-free. As illustrated in \prettyref{fig:vacant_inter_cell}, if robot $\POS_i$ is to be moved to a vacant position $\POS_j$ along an inter-cell edge, then we can rotate $\POS_i$ (along with the two other robots) clockwise and $\POS_j$ counterclockwise until the line-segment $\POS_i-\POS_j$ is orthogonal to $\dee$ and the convex hull of $C(\POS_i)$ and $C(\POS_j)$ does not contain other agents so the vacant move is collision-free. We prove in \prettyref{appen:proof} that such convex hull always exists between any pair of valid triangles. Finally, we rotate $\POS_{i,j}$ reversely to undo extra changes. 
\section{\label{sec:meshopt}Mesh Optimization}
In this section, we present a variant of mesh optimization algorithm \cite{hoppe1993mesh,Hu18} to search for $\DGRAPH$ that maximizes a user given metric $\METRIC(\DGRAPH,\GRAPH)$. Afterwards, a corresponding $\GRAPH$ can be extracted using \prettyref{alg:graph2mesh} to solve for MPP problems.

\subsection{Metric Function}
Our metric consists of 4 terms $\METRIC_{\#r},\METRIC_{\#rc},\METRIC_{sd},\METRIC_{g}$ that can either be a function of the mesh $\DGRAPH$ or the graph $\GRAPH$. These four terms measure the robot packing density as well as the area covered by the robots in $\WORK$. Our first term, $\METRIC_{\#r}$ is defined as: $\METRIC_{\#r}(\GRAPH)\triangleq |\POSS|$, which is the number of vertices contained in $\GRAPH$. This term reflects the number of robots that $\GRAPH$ can hold, but MPP problems might not be feasible for all these robots because $\GRAPH$ might not be connected. $\METRIC_{\#r}$ is a heuristic that guides our algorithm to find more valid cells at an early stage and then try to connect them. Our second term, $\METRIC_{\#rc}$, is the number of robots in the largest connected component of $\GRAPH$:
\begin{align*}
\METRIC_{\#rc}(\GRAPH)\triangleq 
\fmax{\footnotesize{\GRAPH'\subseteq\GRAPH\;\text{connected}}}
(\METRIC_{\#r}(\GRAPH')),
\end{align*}
which is the maximal number of robots for which MPP problems are feasible. Note that $\METRIC_{\#r,\#rc}$ are discrete, non-differentiable terms that are related to both the topology and geometry of $\DGRAPH$. To increase $\METRIC_{\#r,\#rc}$, we need to update the shape of triangles in $\DGRAPH$ and also update their connectivity so that they are valid. Our third term is the agent packing density averaged over the valid cells:
\begin{align*}
\METRIC_{sd}(\GRAPH)\triangleq\frac{\pi r^2 \METRIC_{\#r}(\GRAPH)}{\sum_{\text{valid cell in } \DGRAPH}|\text{cell}|}.
\end{align*}
$\METRIC_{sd}$ is a heuristic guidance term that biases our algorithm towards denser robot packing, when only a subset of the workspace is covered by the robot. Our last metric term $\METRIC_{g}$ is the 2D AMIPS energy \cite{10.1145/2766938} using the smallest regular triangle satisfying condition 2 as the target shape. $\METRIC_{g}$ is differentiable, purely geometric, and related to $\DGRAPH$ only.

\begin{algorithm}[ht]
\begin{small}
\caption{\label{alg:greedy} Optimize $\METRIC$ with respect to $\DGRAPH$.}
\begin{algorithmic}[1]
\State \TE{Input:} Initial $\DGRAPH$
\State Compute $\METRIC\gets\METRIC(\DGRAPH,\MAPPING(\DGRAPH))$
\State Set terminate$\gets$False
\For{pass$\gets1,2$}
\While{Not terminate}
\State Set terminate$\gets$True
\If{Collapsing edges improves $\METRIC_{\#r}+w\METRIC_{\#rc}$}
\State Set terminate$\gets$False
\EndIf
\If{pass$=1$}
\State Splitting edges (i.e. $|\dee|>1.3|\dee^*|$)
\If{Splitting improves $\METRIC_{\#r}+w\METRIC_{\#rc}$}
\State Set terminate$\gets$False
\EndIf
\EndIf
\State Flip edges to reduce $\METRIC_{g}$
\If{Flipping improves $\METRIC_{\#r}+w\METRIC_{\#rc}$}
\State Set terminate$\gets$False
\EndIf
\State Smooth vertices to reduce $\METRIC_{g}$
\If{Smoothing improves $\METRIC_{\#r}+w\METRIC_{\#rc}$}
\State Set terminate$\gets$False
\EndIf
\State LocalOpt. vertices to improve $\METRIC_{sd}$
\If{LocalOpt. improves $\METRIC_{\#r}+w\METRIC_{\#rc}$}
\State Set terminate$\gets$False
\EndIf
\If{GlobalOpt. improves $\METRIC_{\#r}+w\METRIC_{\#rc}$}
\State Get terminate$\gets$False
\EndIf
\EndWhile
\EndFor
\State Return $\MAPPING(\DGRAPH)$
\end{algorithmic}
\end{small}
\end{algorithm}

\subsection{Greedy Maximization of Metric}
\begin{figure}[ht]
\centering
\includegraphics[width=0.99\linewidth]{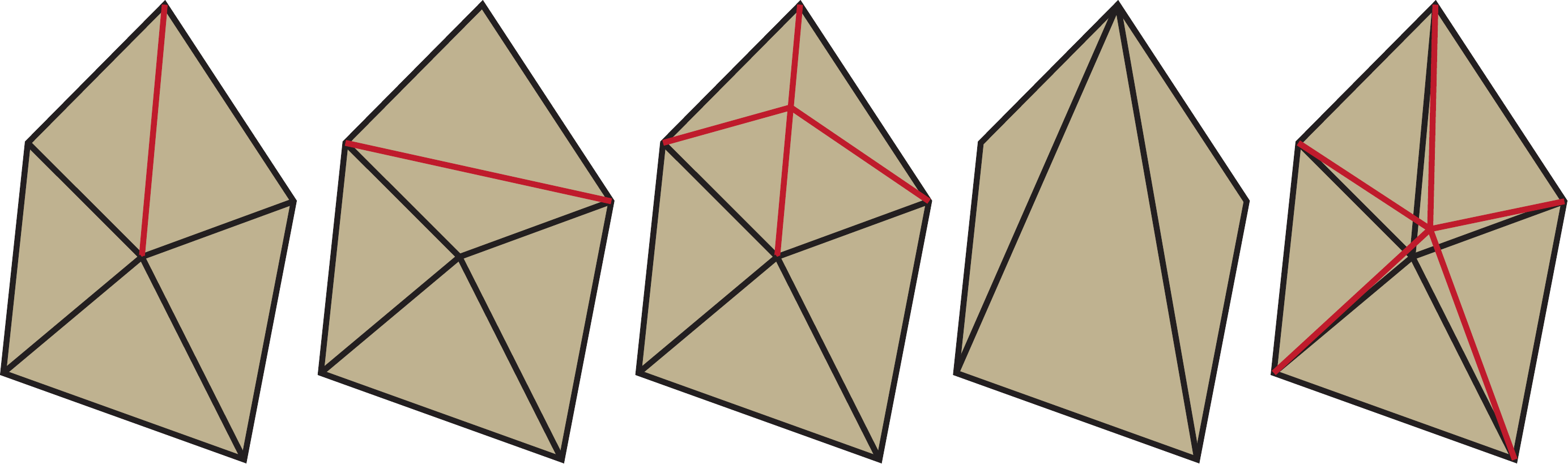}
\put(-240,-5){(a)}
\put(-190,-5){(b)}
\put(-140,-5){(c)}
\put(-90 ,-5){(d)}
\put(-40 ,-5){(e)}
\caption{\small{\label{fig:remeshing} The first 4 remeshing operators inherited from \cite{hoppe1993mesh}. (a): original mesh, (b): edge-flip, (c): edge-split, (d): edge-collapse, (e): vertex-smoothing (part of mesh modified by the operator in red).}}
\vspace{-7px}
\end{figure}
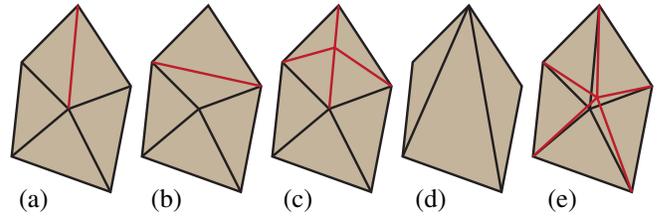

The greedy algorithm is summarized in \prettyref{alg:greedy}, which exhaustively tries one of the re-meshing operators: edge-flip, edge-split, edge-collapse, vertex-smoothing, local-optimization, and global-optimization. Our re-meshing operators, except for global-optimization, are all local and only affect a small neighborhood of cells. The first 4 re-meshing operators are inherited from \cite{hoppe1993mesh,Hu18}, as illustrated in \prettyref{fig:remeshing}. Edge-flip is guided by $\METRIC_{g}$ and is mostly applied to obtuse triangles and turns them into acute triangles. Edge-split and edge-collapse are used to remove tiny edges and split long edges in $\DGRAPH$, respectively. Finally, vertex-smoothing locally optimizes $\METRIC_{g}$ with respect to the one-ring neighborhood of a single vertex. The four operators combined can turn a low-quality mesh into one with nice element shapes.

Our algorithm uses two passes. In the first pass, we allow the algorithm to explore a larger search space by allowing both edge-collapse and edge-split. To ensure the convergence of the first pass, we avoid the cases where consecutive edge-flip and edge-split operators happen for a same edge repeatedly. Allowing edge-split will create more triangles and potentially lead to denser packing of robots. In practice, we split a $\dee$ when its length is larger than $1.3\times$ of the optimal length $\dee^*$. Here the optimal length is the edge length of the smallest regular triangles satisfying condition 2, which equals to: $|\dee^*|=(2\sqrt{3}+4)r$. In the second pass, we are more conservative and disallow edge-split. This pass can be considered as post-processing, merging too small triangles and simplifying the robot layout. Within each pass, we check every possible operator and only accept operators when the weighted metric $\METRIC_{\#r}+w\METRIC_{\#rc}$ monotonically increases and we set $w=10$. We also reject operators that violate our assumption on $\DGRAPH$ being a cell decomposition, i.e. introducing non-manifold connection and flipped cells. The non-manifoldness of $\DGRAPH$ can happen only in edge-collapse and can be avoided by the link condition check~\cite{Dey98topologypreserving}. A flipped cell has a negative area which can be checked by the following constraint:
\begin{align}
\label{eq:flip}
(\DPOS_2-\DPOS_1)\times(\DPOS_3-\DPOS_1)\geq0.
\end{align}

\begin{figure*}[ht]
\centering
\includegraphics[width=0.95\linewidth]{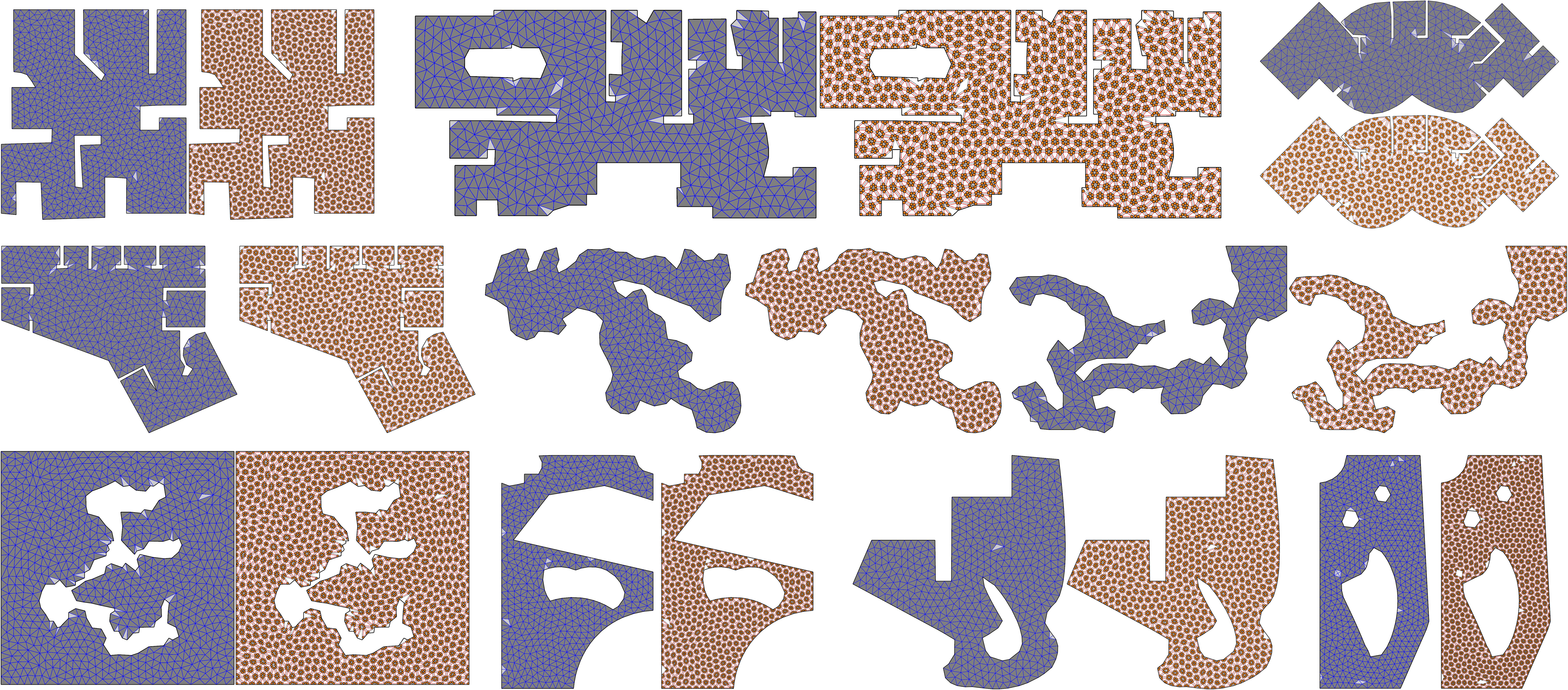}
\caption{\footnotesize{\label{fig:results} Gallery of testing workspaces. For each example, left is our optimized triangulation where cells colored in light gray are invalid, right is the converted graph for MPP problems from valid cells and their connections.}}
\vspace{-5px}
\end{figure*}

\subsection{Local- and Global-Optimization Operators}
Aside from the four remeshing operators inherited from \cite{hoppe1993mesh,Hu18}, we introduce one local- and one global-optimization operators. These operators use the primal-dual interior point method to optimize the geometric shape of the cells, taking condition 2 as hard constraints. At the same time, we maximize $\METRIC_{sd}$ by minimizing the area of valid cells, leading to a denser robot packing. Specifically, for a cell in $\DGRAPH$ with vertices $\DPOS_{1,2,3}$, we solve the following problem in the local-optimization operator:
\begin{equation}
\begin{aligned}
\label{eq:opt}
\argmin{\DPOS_{1,2,3}}&\sum_{\text{valid cell in } \DGRAPH}|\text{cell}|    \\
\E{s.t.}\quad
&\prettyref{eq:equi_cond2}\quad \forall\text{valid cell in } \DGRAPH \\
&\prettyref{eq:flip}\quad \forall\text{cell in } \DGRAPH,
\end{aligned}
\end{equation}
where we require that the entire mesh to be a cell decomposition, so we add \prettyref{eq:flip} for all cells. We also require that those originally valid cells stay valid, so we add \prettyref{eq:equi_cond2} for all valid cells. This optimization can be solved efficiently because it is local. Since we only treat the 3 vertices of a single cell in $\DGRAPH$ as decision variables, most of the terms in the objective function and constraints are outside the 1-ring neighborhood of $\DPOS_{1,2,3}$ and not influenced by the decision variables, which can be omitted. Note that \prettyref{eq:equi_cond2} is expressed in terms of the corner points $\POS_{1,2,3}$, instead of mesh vertices $\DPOS_{1,2,3}$. When the optimizer requires partial derivatives of \prettyref{eq:equi_cond2} against $\DPOS_{1,2,3}$, we use the chain rule on the relationship \prettyref{eq:corner}. 

The global-optimization operator is very similar to the local-optimization operator, which also takes the form of \prettyref{eq:opt}, but we set all the variables $\DPOS$ as decision variables. The global-optimization is more expensive than all other operators, but we found that this last operator can significantly improve robot packing density and it is more efficient than applying the local-optimization operator to each of the cell.

\section{\label{sec:results}Results and Analysis}
We implement our algorithm and conduct experiments on a single desktop machine with Intel i7-9700 CPU. The input to our algorithm is a vector image in the svg format representing $\WORK$. Our main \prettyref{alg:greedy} starts from an initial guess of $\DGRAPH$, which can be generated using Triangle \cite{triangle}. Our mesh optimization algorithm can nicely turn a triangle mesh (left of the inset), with an arbitrary initialization that has only few triangles satisfying the graph embedding constraints, into a regular triangle mesh (right of the inset) with most of its triangles being valid graph nodes for MPP problems.
\begin{wrapfigure}{r}{0.25\textwidth}
\vspace{-5px}
\includegraphics[width=0.25\textwidth]{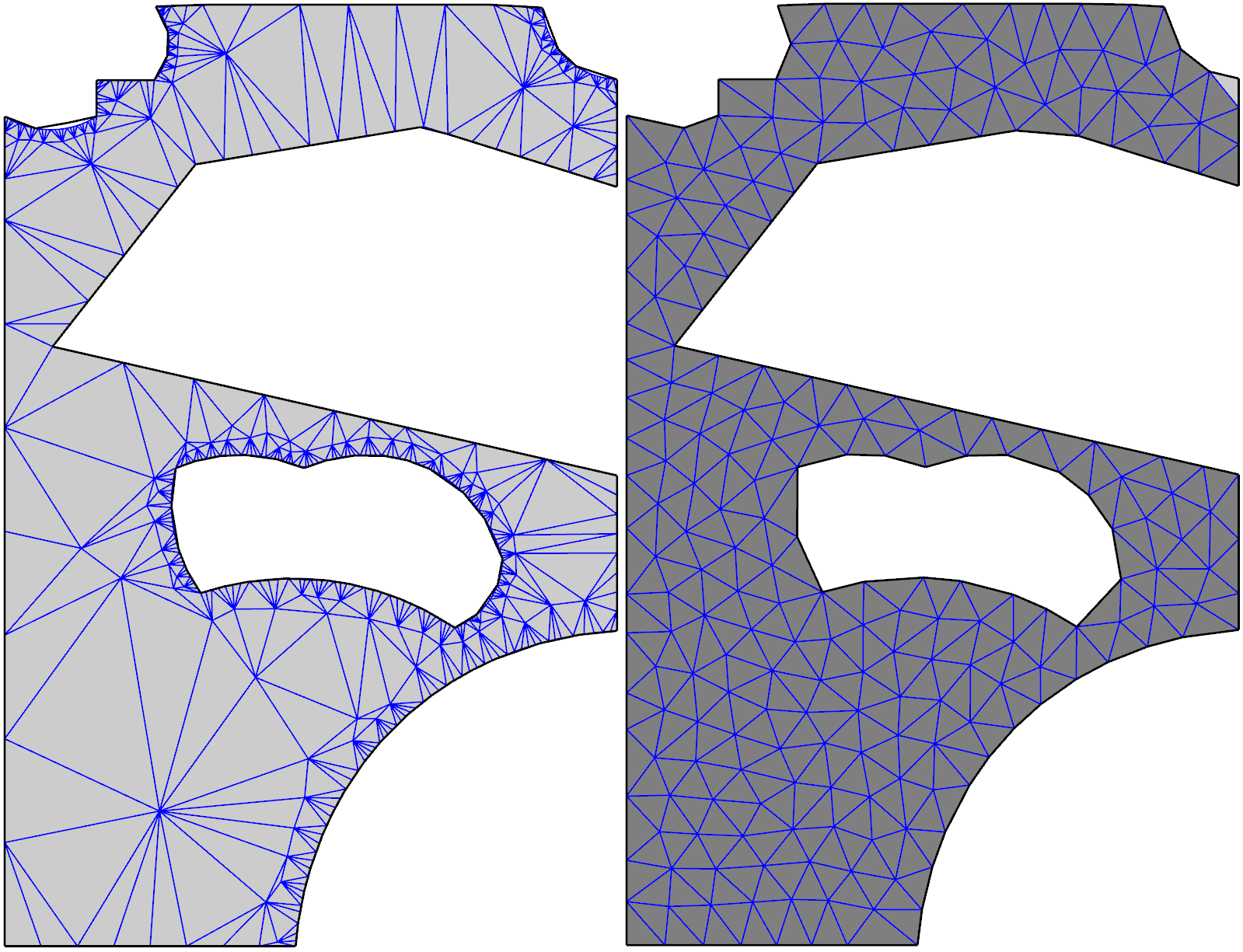}
\vspace{-15px}
\end{wrapfigure}
We evaluate our algorithm on a set of complex workspaces and we summarize the statistics of $\METRIC_{\#r}$, $\METRIC_{\#rc}$, robot density $\mathcal{D}$, free space coverage (the ratio between the area of all valid cells and the area of $\WORK$), and computational time of the mesh optimization in \prettyref{fig:results} and \prettyref{table:results}.

\begin{table}[ht]
\setlength{\tabcolsep}{1pt}
\begin{center}
\footnotesize
\begin{tabular}{lcccccccccc}
\toprule
\backslashbox[15mm]{{\scriptsize Measures}}{{\scriptsize Models}}
& 1 & 2 & 3 & 4 & 5 & 6 & 7 & 8 & 9 & 10\\\hline
$\METRIC_{\#r}$& 
       1489 & 
       1061 & 
       782 & 
       1021 & 
       756 & 
       569 & 
       1515 & 
       877 & 
       981 & 
       1104 \\
$\METRIC_{\#rc}$& 
       1489 & 
       1061 & 
       782 & 
       1021 & 
       756 & 
       569 & 
       1515 & 
       877 & 
       981 & 
       1104 \\
$\mathcal{D}$&
       0.31&
       0.29&
       0.30&
       0.30&
       0.29&
       0.28&
       0.30&
       0.31&
       0.32&
       0.33\\
Coverage&
       0.995&
       0.979&
       0.983&
       0.984&
       1.000&
       0.994&
       0.985&
       0.993&
       0.998&
       0.990\\
time&
       14133&
       2598&
       4439&
       5866&
       2793&
       3241&
       6155&
       2539&
       7836&
       9066\\
       \hline
\end{tabular}
\end{center}
\vspace{-5px}
\caption{\footnotesize{\label{table:results}Statistics of $\METRIC_{\#r}$, $\METRIC_{\#rc}$, robot density $\mathcal{D}$, free space coverage, and time (s) measurements for the graph generation of the workspaces listed in \prettyref{fig:results} (indexed from top to bottom, left to right).}}
\vspace{-10px}
\end{table}

\begin{figure}[ht]
\centering
\includegraphics[width=\linewidth]{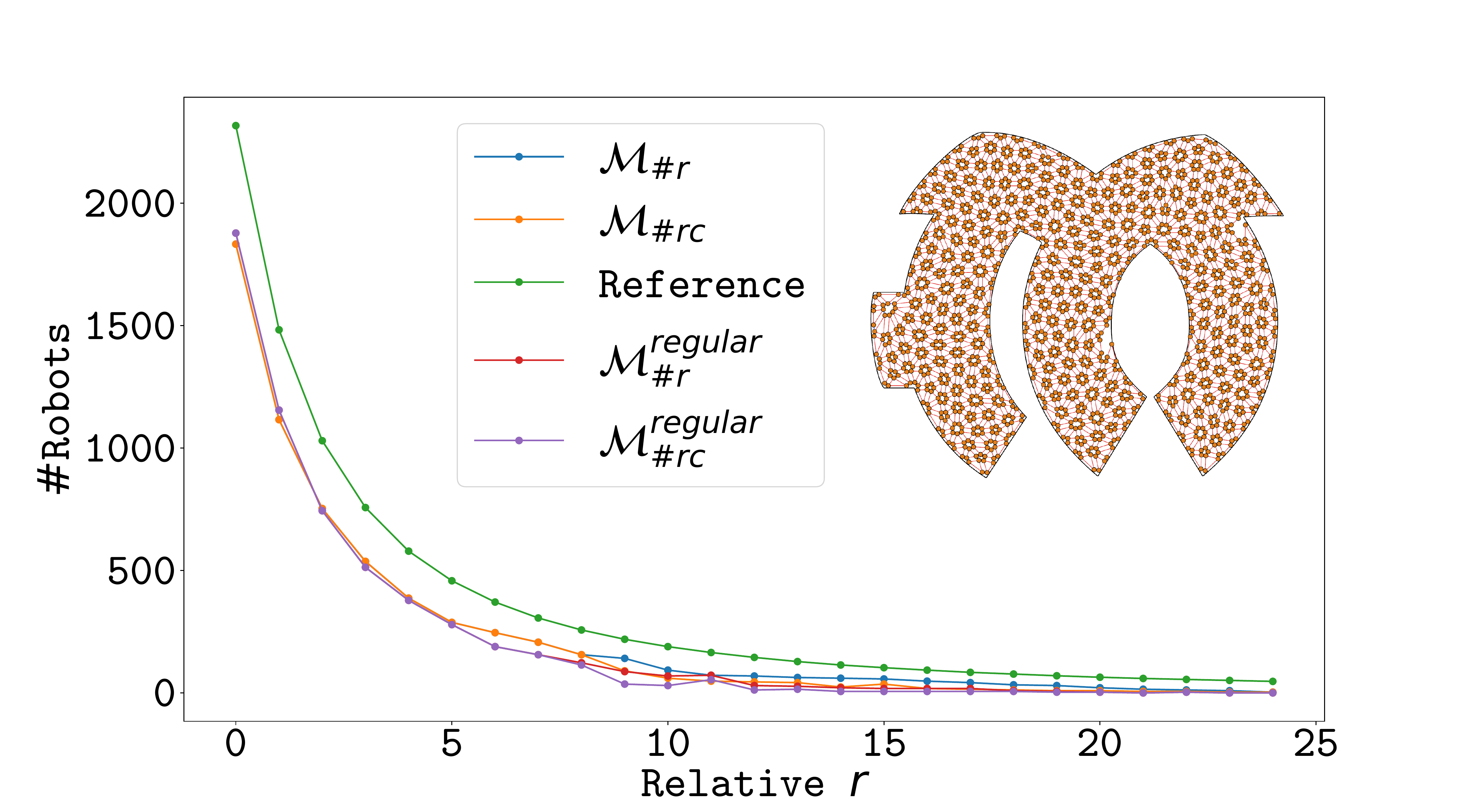}
\caption{\small{\label{fig:numrobotchange} By varying $r$, we plot $\METRIC_{\#r}$ and $\METRIC_{\#rc}$ produced by our boundary aligned approach, $\METRIC^{regular}_{\#r}$ and $\METRIC^{regular}_{\#rc}$ by the simple regular-pattern baseline, and compare them with the ideal reference $3|\WORK|/A^*$.}}
\vspace{-10px}
\end{figure}

We also compare our method with the regular-pattern baseline, illustrated in \prettyref{fig:irregular} and \prettyref{fig:numrobotchange}. Using our arrangement of robots in \prettyref{sec:graph2mesh}, we can use regular triangles with the optimal area $A^*$ to cover $\WORK$ and then remove triangles that are outside $\WORK$. As shown in \prettyref{fig:numrobotchange}, the regular pattern does lead to higher $\METRIC_{\#r}$ and $\METRIC_{\#rc}$ when the size of a robot is small. But, as the robot size gets larger, our method can produce a graph embedding with a higher number of robots (\prettyref{fig:irregular} and \prettyref{fig:numrobotchange}). Moreover, regular pattern method introduces undesirable zig-zag boundary coverage (see \prettyref{fig:irregular} for a visual comparison on a simple workspace).

We also profile the influence of different robot radius. In \prettyref{fig:numrobotchange}, we change the radius $r$, and compare the curve with the ideal curve $3|\WORK|/A^*$ by plotting the change of $\METRIC_{\#r}$ and $\METRIC_{\#rc}$ against $r$. Our method closely matches the ideal reference curve and the performance is comparable to regular-pattern baseline with different $r$.

\begin{figure}[ht]
\centering
\includegraphics[width=\linewidth]{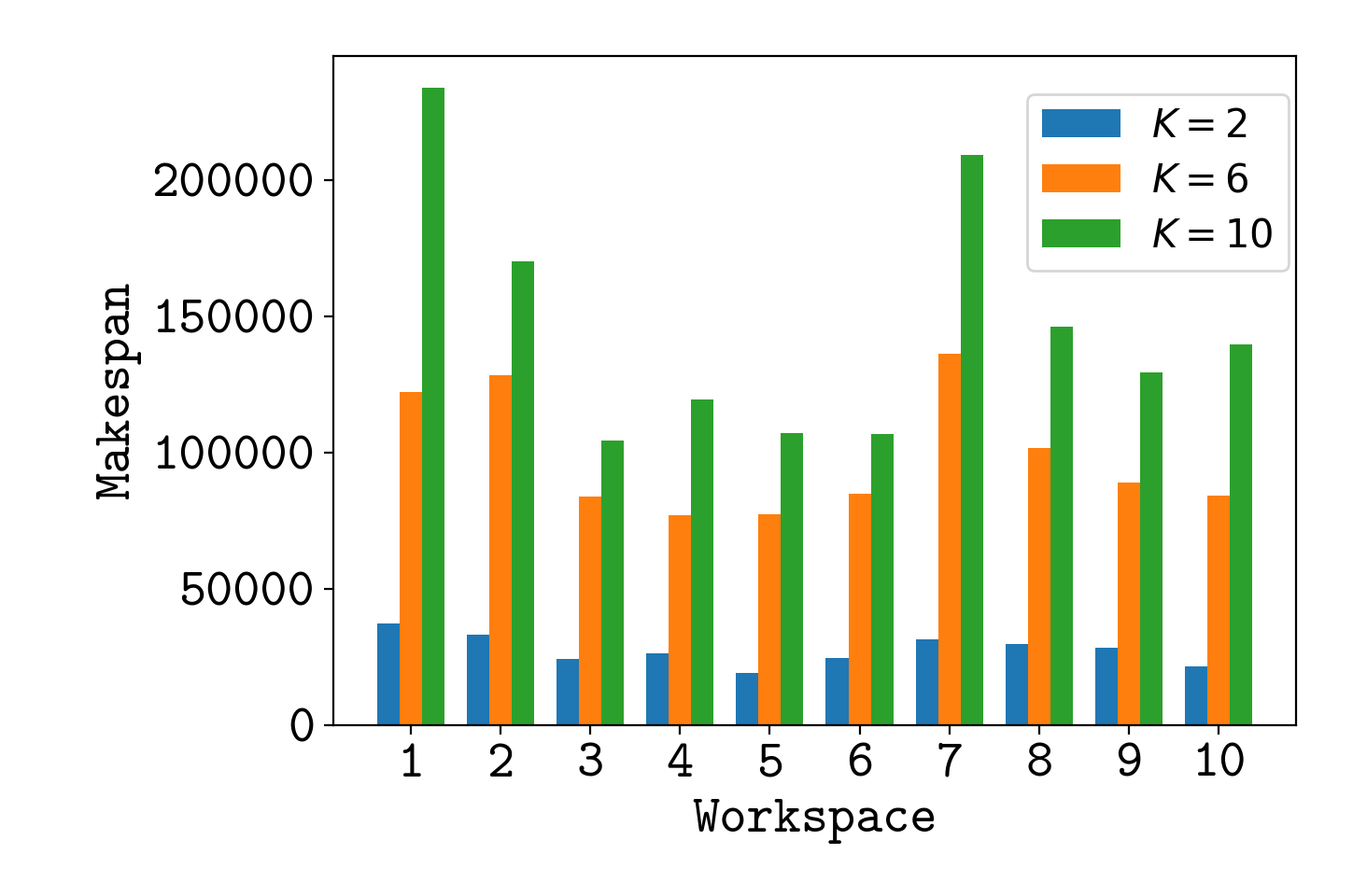}
\caption{\small{\label{fig:makespan} We summarize the average makespan for $10$ environments under three different $K$. For most environments, the makespan reduces linearly with $K$.}}
\vspace{-10px}
\end{figure}

On the optimized $\DGRAPH$ and corresponding $\GRAPH$, we use our parallel MPP scheduler \prettyref{alg:spacetime} to solve $5$ randomized MPP problems for each of the $10$ environments listed in \prettyref{fig:results} and \prettyref{table:results}. We profile the rate of acceleration using more and more vacant vertices (determined by the parameter $K$). We summarize the average makespan for each environment for three different $K$s in \prettyref{fig:makespan}. We observe greater makespan reductions as $K$ goes smaller that will consider more agents as vacant.
\section{\label{sec:conclusion}Conclusion and Limitations}
We present a method to solve graph-restricted MPP problems in complex environments. Our method uses a special graph topology to ensure MPP feasibility. Using a divide-and-conquer algorithm with space-time scheduling, we can parallelize the robot motions and greedily improve the makespan. The special graph topology can be further identified with triangular meshes, which in turn is optimized via mesh optimization and embedded into arbitrarily complex environments. In the future, we plan to extend our method to continuous robot motions under general dynamic constraints.
\AtNextBibliography{\tiny}
\printbibliography
\section{Appendix: Proof of Vacant Move\label{appen:proof}}
\begin{lemma}
Assuming condition 2 of \prettyref{sec:graph2mesh}, for a pair of positions $\POS_{i,j}$ in different triangles sharing an edge $\dee$, $\POS_i$ and $\POS_j$ can be rotated such that the convex hull of $C(\POS_i)$ and $C(\POS_j)$ does not contain other agents.
\end{lemma}
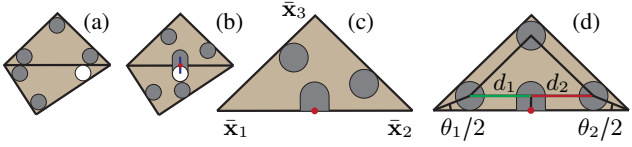
\begin{figure}[ht]
\centering
\includegraphics[width=0.99\linewidth]{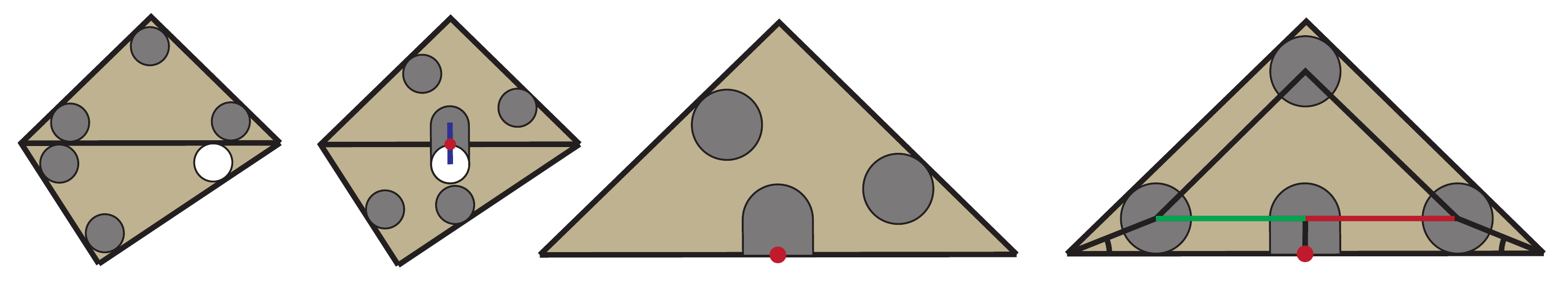}
\put(-210,35){\small{(a)}}
\put(-160,35){\small{(b)}}
\put(-110,35){\small{(c)}}
\put(-25,35){\small{(d)}}
\put(-157,-5){\small{$\DPOS_1$}}
\put(-95 ,-5){\small{$\DPOS_2$}}
\put(-135,40){\small{$\DPOS_3$}}
\put(-55,12){\small{$d_1$}}
\put(-35,12){\small{$d_2$}}
\put(-75,-5){\small{${\theta_1}/{2}$}}
\put(-23,-5){\small{${\theta_2}/{2}$}}
\caption{\small{\label{fig:vacant_inter_proof} The red dot is the center of the shared edge, $d_{1,2}$ are the length of the green,red edge respectively, and $\theta_{1,2}/2$ are the two half angles in (d).}}
\vspace{-15px}
\end{figure}
\begin{proof}
As illustrated in \prettyref{fig:vacant_inter_proof} (ab), we have to align $\POS_i$ and $\POS_j$ by rotating $\POS_i$ clockwise and $\POS_j$ counterclockwise. We choose to always rotate until the edge between $\POS_i$ and $\POS_j$ passes through the center of the shared edge: $(\DPOS_1+\DPOS_2)/2$ in \prettyref{fig:vacant_inter_proof} (c). Therefore, we only need to consider one triangle and show that the half capsule lies inside the triangle (\prettyref{fig:vacant_inter_proof} (c)), for which an equivalent condition is $d_1\geq r$ and $d_2\geq r$ in \prettyref{fig:vacant_inter_proof} (d) for all possible choices of $\theta_{1,2}$.

We first derive closed-form expressions for $d_{1,2}$. Note that at least one of the two angles $\theta_{1,2}$ are acute and, without loss of generality, we assume that $\theta_2$ is acute. We further denote $\eta_1\triangleq\tan(\theta_1/2)$ and $\eta_2\triangleq\tan(\theta_2/2)$. Condition 2 implies the following two conditions:
\tiny
\begin{align}
\label{eq:E1}
\|\POS_2(t)-\POS_1(t)\|&\geq2r\quad\forall t\in[0,1]\\
\label{eq:E2}
\|\POS_2(t)-\POS_3(t)\|&\geq2r\quad\forall t\in[0,1].
\end{align}
\normalsize
Since $\theta_2$ is an acute angle, it is easy to show that $\|\POS_2(t)-\POS_1(t)\|$ attains its minimal value when $t\in[0,1]$. Therefore, we can eliminate $t$ and derive a lower bound of $d_1$ from \prettyref{eq:E1} (This is lower bound because \prettyref{eq:E2} is omitted). After trigonometric simplification, we can express the lower bound of $d_1-r$ as the following functions of $\eta_{1,2}$:
\tiny
\begin{align*}
&d_1-r\geq(L_1-R_1)r/(2\eta_1^2\eta_2^2) \\
&L_1\triangleq \eta_1\eta_2
\sqrt{4\eta_1^2\eta_2^4+(4\eta_1^3-4\eta_1)\eta_2^3+(1-6\eta_1^2+\eta_1^4)\eta_2^2+(4\eta_1-4\eta_1^3)\eta_2+4\eta_1^2}   \\
&R_1\triangleq (-\eta_1-2\eta_1^2)\eta_2+\eta_1^2\eta_2.
\end{align*}
\normalsize
To show that $d_1$ is always positive, we define $\Delta_1\triangleq L_1^2-R_1^2$ and show that $\Delta_1$ is always larger than zero on the following semi-algebraic set:
\begin{align*}
\mathcal{K}_1\triangleq\{\TWO{\eta_1}{\eta_2}|\eta_2\in(0,1),\eta_1>0,\eta_1\eta_2<1\}.
\end{align*}
Note that $\Delta_1$ is a polynomial function and verifying the positivity of a polynomial function on a semi-algebraic set can be performed analytically using Gr\"obner basis \cite{hagglof1995computing}. We use the computational algebraic software Maxima \cite{li2008maxima} to verify that the global minimum of $\Delta_1$ on $\mathcal{K}_1$ is zero. Therefore, we conclude that $d_1\geq r$ for any valid triangles. Similarly, we can derive the lower bound of $d_2$ resulting in a similar formula:
\tiny
\begin{align*}
&d_2-r\geq(L_2-R_2)r/(2\eta_1^2\eta_2^2) \\
&L_2\triangleq \eta_1\eta_2
\sqrt{4\eta_1^2\eta_2^4+(4\eta_1^3-4\eta_1)\eta_2^3+(1-6\eta_1^2+\eta_1^4)\eta_2^2+(4\eta_1-4\eta_1^3)\eta_2+4\eta_1^2}   \\
&R_2\triangleq (\eta-2\eta_1^2)\eta_2-\eta_1^2\eta_2.
\end{align*}
\normalsize
If we define $\Delta_2\triangleq L_2^2-R_2^2$ and compute the global minimum of $\Delta_2$ on $\mathcal{K}_1$, the result is negative. Therefore, we need to consider two cases separately. First, if $\theta_1+\theta_2<\pi/2$, then we have the following semi-algebraic set:
\begin{align*}
\mathcal{K}_2\triangleq\{\TWO{\eta_1}{\eta_2}|\eta_2\in(0,1),\eta_1>0,\eta_1+\eta_2+\eta_1\eta_2<1\}.
\end{align*}
The global minimum of $\Delta_2$ on $\mathcal{K}_2$ is zero. Therefore, we conclude that $d_2\geq r$ holds for any valid triangles with $\theta_1+\theta_2<\pi/2$. If $\theta_1+\theta_2>\pi/2$, then the third angle $\pi-(\theta_1+\theta_2)$ is acute and we can show that the left-hand side of \prettyref{eq:E2} attains its minimum when $t\in[0,1]$. As a result, \prettyref{eq:E2} gives us another lower bound for $d_2$. Applying trigonometric simplification and we have:
\tiny
\begin{align*}
&d_2-r\geq(L_3-R_3)r/(2\eta_1^2\eta_2^2) \\
&L_3\triangleq \eta_1\eta_2
\sqrt{\eta_1^2\eta_2^4+(2\eta_1-2\eta_1^3)\eta_2^3+(1+12\eta_1^2+\eta_1^4)\eta_2^2+(2\eta_1^3-2\eta_1)\eta_2+\eta_1^2}   \\
&R_3\triangleq (\eta_1-2\eta_1^2)\eta_2-\eta_1^2\eta_2.
\end{align*}
\normalsize
If we define $\Delta_3\triangleq L_3^2-R_3^2$, we can show that the global minimum of $\Delta_3$ on $\mathcal{K}_1-\mathcal{K}_2$ is zero. As a result, $d_2\geq r$ for any valid triangles with $\theta_1+\theta_2\geq\pi/2$.
\end{proof}
\end{document}